\documentclass[letterpaper, 10 pt, conference]{ieeeconf}  

\IEEEoverridecommandlockouts                              

\usepackage{preamble}

\title{\LARGE \bf
Are Learning-Based Approaches Ready for Real-World Indoor Navigation? A Case for Imitation Learning
}

\author{Nigitha Selvaraj$^{\mathsection}$, Alex Mitrevski$^{\mathsection}$, and Sebastian Houben
\thanks{$^{*}$This work was supported by the b-it foundation} %
\thanks{$^{\dagger}$The authors are with the Autonomous Systems Group, Institute of AI and Autonomous Systems (A$^2$S), Hochschule Bonn-Rhein-Sieg, Sankt Augustin, Germany\newline
        {\tt\scriptsize nigitha.selvaraj@smail.inf.h-brs.de, <aleksandar.mitrevski, sebastian.houben>@h-brs.de}} %
\thanks{$^{\mathsection}$Corresponding author} %
}

\begin{document}

\maketitle
\thispagestyle{empty}
\pagestyle{empty}


\begin{abstract}
    Traditional indoor robot navigation methods provide a reliable solution when adapted to constrained scenarios, but lack flexibility or require manual re-tuning when deployed in more complex settings.
    In contrast, learning-based approaches learn directly from sensor data and environmental interactions, enabling easier adaptability.
    While significant work has been presented in the context of learning navigation policies, learning-based methods are rarely compared to traditional navigation methods directly, which is a problem for their ultimate acceptance in general navigation contexts.
    In this work, we explore the viability of imitation learning (IL) for indoor navigation, using expert (joystick) demonstrations to train various navigation policy networks based on RGB images, LiDAR, and a combination of both, and we compare our IL approach to a traditional potential field-based navigation method.
    We evaluate the approach on a physical mobile robot platform equipped with a 2D LiDAR and a camera in an indoor university environment.
    Our multimodal model demonstrates superior navigation capabilities in most scenarios, but faces challenges in dynamic environments, likely due to limited diversity in the demonstrations.
    Nevertheless, the ability to learn directly from data and generalise across layouts suggests that IL can be a practical navigation approach, and potentially a useful initialisation strategy for subsequent lifelong learning.
\end{abstract}


    \section{INTRODUCTION}
    \label{sec:introduction}

    Indoor robot navigation is a core capability for autonomous systems deployed in structured yet dynamic environments, such as offices, hospitals, and warehouses.
    These settings demand precise perception, decision-making, and execution to navigate safely and efficiently.
    Traditional local path planning algorithms rely on geometric and heuristic rules, which, while effective in predictable scenarios, often struggle with adaptability in more challenging settings \cite{Zhou_2021}.
    
    Recent advancements in machine learning have opened new possibilities for addressing these limitations, as learning-based approaches enable robots to learn navigation policies directly from sensor data, bypassing the need for manual tuning and rule-based definitions \cite{xiao2022_survey,kiran2022,levine2023}.
    Using sensory input and interaction data, these systems have the potential to achieve a higher degree of adaptability and robustness.
    Among these methods, imitation learning (IL) \cite{ravichandar2020} stands out for its simplicity and data efficiency, as it enables robots to acquire navigation behaviors directly from expert demonstrations.
    Nevertheless, deploying learning-based systems in real-world scenarios introduces several challenges, and its practicality, particularly as opposed to traditional model-based navigation methods, remains largely unexplored.

    Our study aims to answer the following questions:
    \begin{itemize}
        \item How does IL compare to a traditional motion planner, in particular a potential field-based method, in terms of navigation success and reliability?
        \item How well can IL leverage different sensor modalities, such as RGB and LiDAR, to improve navigation performance in real-world environments?
    \end{itemize}
    The concrete objective is to develop a learning-based navigation framework for mobile robots to navigate real-world indoor environments using IL, while investigating the above questions to assess the practical potential of IL for real-world deployment.

    For this, we propose a learning-based navigation framework centered on imitation learning (IL), where expert demonstrations are collected using a joystick-operated mobile robot across a variety of indoor layouts.
    IL models are then trained with three distinct sensory input configurations: RGB images, 2D LiDAR scans, and a fusion of both modalities.
    The trained policies are deployed on a mobile robot (a KELO ROBILE platform\footnote{\url{https://shop.kelo-robotics.com/product/standard-configuration/}}), and evaluated in a real-world indoor university environment.
    To assess the navigation effectiveness and generalisation capabilities, we compare the performance of the learning-based method against a classical potential field planner \cite{khatib1985}.
    The results show that, with the laser-based model, the robot showed an overall of 70\% success rate, while the multi-modal model combining laser and image data showed 86.7\% success rate in static environments, performing better than classical potential field planner in scenarios with confined space.\footnote{Accompanying video: \url{https://youtu.be/TCvOwCVyuNo}}

    \section{RELATED WORK}
    \label{sec:related_work}

    \begin{table*}[t]
        \caption{Comparison of our approach with other learning-based navigation methods}
        \label{tab:soa_table}
        \begin{tabular}{M{0.23\linewidth} M{0.14\linewidth} M{0.185\linewidth} M{0.1\linewidth} M{0.1\linewidth} M{0.1\linewidth}}
            \hline
            \cellcolor{gray!10!white}Approach & \cellcolor{gray!10!white}Sensor modalities & \cellcolor{gray!10!white}Environment & \cellcolor{gray!10!white}Expert data & \cellcolor{gray!10!white}Real-world experiments & \cellcolor{gray!10!white}Baseline comparison \\\hline
            Map-based DRL \cite{Chen2020} & 2D Laser scanner & Indoor room with boxes as obstacles & \xmark & \cmark & \xmark \\\hline
            RL for pedestrian environments \cite{PerezD’Arpino2021} & 1D LiDAR & Indoor pedestrian environments & \xmark & \xmark & \cmark \\\hline
            Target-driven visual navigation using DRL \cite{Zhu2017} & RGB Camera & Indoor scenes such as living room & \xmark & \cmark & \xmark \\\hline
            Multimodal DRL with auxiliary task for obstacle avoidance \cite{Song2021} & 2D laser range, Depth image & Indoor & \xmark & \cmark & \xmark \\\hline
            Self-supervised DRL with generalised computation graphs \cite{Kahn2018} & Gray scale image & Cluttered indoor environment & \xmark & \cmark & \xmark \\\hline
            Self-supervised learning \cite{Kahn2021} & RGB Camera, LiDAR, IMU & Outdoor environment - urban and off-road & \xmark & \cmark & \xmark \\\hline
            DRL with self-state attention and sensor fusion \cite{Han2022} & RGB Camera and 2D LiDAR & Multiple indoor scenes & \xmark & \cmark & \xmark \\\hline
            RL-based dynamic obstacle avoidance and integration of path planning \cite{Choi2021} & 2D LiDAR & Multirobot indoor scenarios & \xmark & \cmark & \cmark \\\hline
            Pretraining with IL and RL finetuning \cite{Ramrakhya2023} & RGB Camera, GPS, Compass & Indoor environment with static obstacles & \cmark & \xmark & \xmark \\\hline
            \textbf{Ours} & \textbf{2D laser scan, RGB images} & \textbf{Indoor - university hallway} & \cmark & \cmark & \cmark \\\hline
        \end{tabular}
    \end{table*}

    In the field of (indoor) robotic navigation, techniques are continuously refined to achieve more efficient and reliable path and motion planning.
    Recent studies in this area have introduced significant learning-based approaches that improve the adaptability and precision of robotic navigation in complex indoor environments.

    Supervised learning is commonly used for tasks such as regression (predicting steering angles or speed) \cite{Das_2022} and classification (detecting obstacles and collisions) \cite{Becker_2019}.
    However, purely supervised approaches rely heavily on the quality and diversity of the training data and require manual labeling.

    To reduce the data dependency, self-supervised learning has emerged as an alternative.
    In self-supervised methods \cite{Kahn2018}, the robot generates its own training signals from interactions with the environment, reducing or even eliminating the need for manually labeled data.
    BADGR \cite{Kahn2021} is one such method, in which a mobile robot system autonomously collects data of collision events to train predictive models; however, such methods still require the robot to encounter failures, which can be risky in real deployments.
 
    Deep reinforcement learning (DRL) is another paradigm for enabling robots to autonomously learn navigation policies directly from sensor data \cite{Zhu_2021,xiao2022_survey}.
    One line of work focuses on using visual inputs (in particular, camera images) as inputs to DRL methods to navigate towards specified visual targets \cite{Zhu2017}.
    Additionally, multimodal deep reinforcement learning approaches have been explored to improve obstacle avoidance capabilities by leveraging different sensory inputs, such as vision and depth information, resulting in more robust navigation performance \cite{Song2021}.
    Methods incorporating self-state attention and sensor fusion have enhanced collision avoidance by effectively using sensory data \cite{Han2022}.
    Yet, real-world training remains challenging due to safety concerns, unstable learning, and high sample demands.

    Imitation Learning (IL) addresses many of these challenges by learning directly from expert demonstrations, enabling safe and efficient policy learning without reward design \cite{Pfeiffer_2017}.
    For instance, IL has been used to train a motion planner based on a convolutional neural network (CNN) and a long short-term memory (LSTM) network \cite{Hoshino_2022}.
    The reliance on high-quality expert demonstrations means that errors or biases in the demonstrations can propagate into the learned policy, requiring proper evaluation of IL approach before real-world deployment.
    Our work attempts to address this aspect by performing a comparison of a learning-based navigation approach with a traditional, potential field-based navigation method.

    Tab. \ref{tab:soa_table} provides a comparison of our approach with other related approaches for (indoor) navigation based on various factors, including the sensory modalities used, the types of environments in which they are applied, the use of expert data, their real-world applicability, and whether they have been benchmarked against traditional navigation baselines.

    \section{METHODOLOGY}
    \label{sec:methodology}

    The objective of this study is to investigate the feasibility of setting up a learning-based robot navigation system for indoor environments and to perform a comparison of such a system with a classical navigation method.
    For this purpose, we train multiple neural network-based goal-conditioned navigation policies that use LiDAR data, image data, or fused data from both modalities; concretely, we collect expert demonstrations and use imitation learning (IL) to learn the policies, which we then compare against a potential field method.
    We use network-based policies due to their flexibility in processing different input modalities; this is particularly important because we use image inputs in a subset of the policies.
    In this section, we describe the process of collecting policy learning data, the policy networks, and the IL process; the experimental evaluation and comparison is then presented in the next section.

    \subsection{Imitation Learning}

    The IL phase teaches a robot to imitate expert demonstrations by learning from sensor data and corresponding motion commands.
    We concretely use behavioral cloning (BC) as an IL approach, which treats the navigation problem as a supervised learning task, namely the model is trained to directly predict an expert's motion commands (linear and angular velocities) based on sensory inputs, such as RGB images, laser scans, and goal positions.
    The training objective is to minimise the error between the predicted and expert (ground-truth) commands; we use the mean squared error (MSE) loss function for this purpose:
    \begin{equation}
        \label{eq:il_loss}
        \mathcal{L} = \frac{1}{N} \sum_{i=1}^{N} \lVert \hat{\vec{a}}_i - \vec{a}_i \rVert^2
    \end{equation}
    Here, $\mathcal{L}$ is the total loss, $N$ is the total number of data points, $\vec{a}_i$ is the expert motion command (for planar navigation, this is a 3D vector with linear (2D) and angular (1D) velocity) for the $i$-th data point, and $\hat{\vec{a}}_i$ is the predicted motion command for the $i$-th data point.
    The benefit of BC is that it does not require interaction with the environment during training, namely it allows for fast and stable learning using only offline expert demonstrations.

    \subsection{Data Collection and Preprocessing}
    \label{sec:methodology:data_collection}

\begin{figure}[t]
        \centering
        \includegraphics[width=0.7\linewidth]{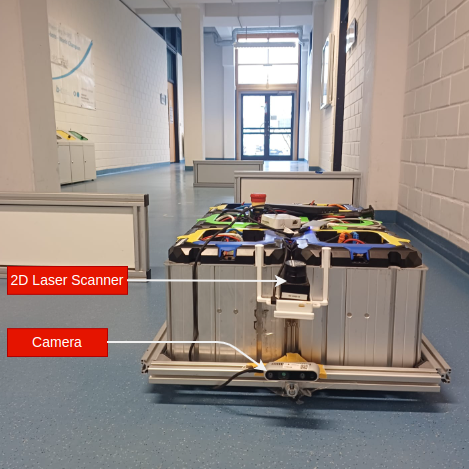}
        \caption{Mobile robot used in this study}
        \label{fig:robot}
    \end{figure}

    We collected expert demonstrations to pretrain the policy networks using IL.\footnote{For the purposes of this study, the demonstrations were collected by the researchers; no external participants were involved in the process.}
    A KELO ROBILE, an autonomous mobile robot with omnidirectional wheels, was used to navigate in an indoor environment (a university hallway set up with obstacles), as illustrated in Fig. \ref{fig:robot}.
    The robot was manually operated using a joystick across 19 trajectories of straight-line and curvilinear motion, each approximately 3 meters long, to collect data from various sensor modalities.
    For the data collection, the operator controlled the robot to perform goal-directed navigation at different speeds (generally low to ensure data stability), avoiding obstacles and reaching targets.
    The collected data consisted of:
    \begin{itemize}
        \item \emph{RGB images} from an Intel Realsense d435i camera
        \item \emph{Laser scans} from a Hokuyo UST-10LX 2D scanner
        \item \emph{Odometry data}, consisting of the robot's position and velocity
        \item \emph{Transform (TF) data} to map positional transformations and orientations with respect to the robot's base frame
    \end{itemize}

    To create a synchronised dataset suitable for IL, the sensor data was aligned and processed as follows:
    \begin{itemize}
        \item The data streams were downsampled to reduce computational overhead, and were synchronised based on the lowest data frequency ($10Hz$ in our setup).
        \item Data points where the robot remained stationary for prolonged durations were filtered out.
        \item From the robot's future trajectory, multiple poses that are between $0.2m$ and $3m$ ahead were selected as goals; this ensures a good mix of short-range and long-range goals for the robot to learn. Using the transform data, \emph{the goal is defined as the distance and orientation with respect to the robot's base frame}.
        \item Invalid laser scan values were replaced with a high numerical value (to indicate an infinite measurement).
        \item Random variations in brightness, contrast, and saturation were introduced to augment the image data and thus enhance the dataset diversity.
        \item Outlier detection was not performed so that the policy network can learn the real data distribution.
    \end{itemize}
    The final dataset consists of laser scan data, augmented images, relative goal poses in terms of the desired direction and orientation, and motion commands. 

    \subsection{Policy Networks}

\begin{figure}[t]
        \centering
        \begin{subfigure}[b]{0.45\linewidth}
            \centering
            \includegraphics[width=\linewidth]{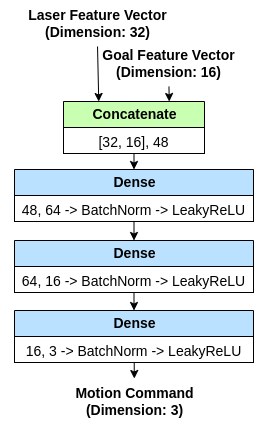}
            \caption{Model with laser and goal as inputs}
            \label{fig:laser_goal_model}
        \end{subfigure}
        \hfill
        \begin{subfigure}[b]{0.45\linewidth}
            \centering
            \includegraphics[width=\linewidth]{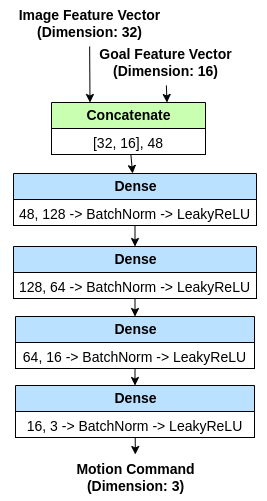}
            \caption{Model with image and goal as inputs}
            \label{fig:image_goal_model}
        \end{subfigure}
        \vfill
        \vspace{0.5cm}
        \begin{subfigure}[b]{0.5\linewidth}
            \centering
            \includegraphics[width=\linewidth]{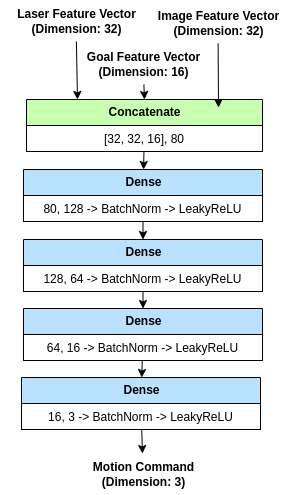}
            \caption{Model with laser, image and goal as inputs}
            \label{fig:laser_image_goal_model}
        \end{subfigure}
        \caption{Architectures of the three policies with different input modalities}
        \label{fig:models}
    \end{figure}

    To investigate different combination of modalities, we developed three distinct policy networks, each tailored to specific input modalities: (i) laser and goal, (ii) image and goal, and (iii) a combination of all three (laser, image, and goal).
    Fig. \ref{fig:models} illustrates the policy network architectures.

\begin{figure}[t]
        \centering
        \begin{subfigure}[b]{0.45\linewidth}
            \centering
            \includegraphics[width=\linewidth]{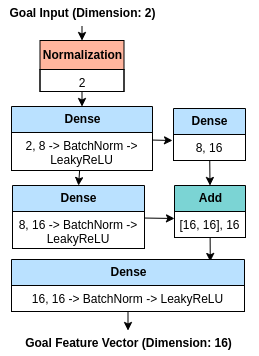}
            \caption{Goal submodel}
            \label{fig:goal_submodel}
        \end{subfigure}
        \hfill
        \begin{subfigure}[b]{0.45\linewidth}
            \centering
            \includegraphics[width=\linewidth]{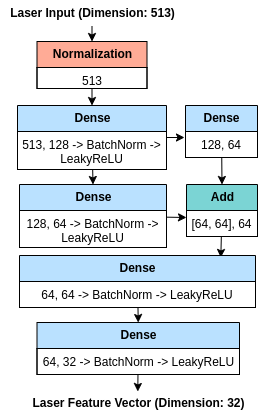}
            \caption{Laser submodel}
            \label{fig:laser_submodel}
        \end{subfigure}
        \vfill
        \vspace{0.5cm}
        \begin{subfigure}[b]{0.9\linewidth}
            \centering
            \includegraphics[width=\linewidth]{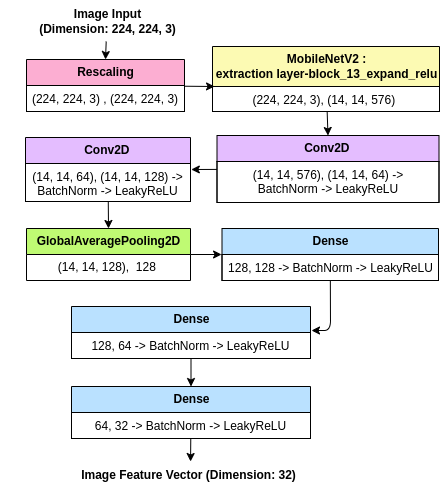}
            \caption{Image submodel}
            \label{fig:image_submodel}
        \end{subfigure}
        \caption{Submodels to extract feature representations of the different modalities}
        \label{fig:sub_models}
    \end{figure}

    The architectures are designed to extract relevant features from each modality and effectively combine them to predict motion commands.
    For feature extraction, dedicated submodels are designed and integrated into the policy architectures:
    \begin{itemize}
        \item \emph{Laser submodel}: The laser scans are normalised (zero mean and unit variance) and processed through dense layers and residual blocks to extract high-level features.
        \item \emph{Goal submodel}: The goal data is normalised (zero mean and unit variance) and processed similarly to the laser model with just one dense layer and one residual block.
        \item \emph{Image submodel}: The RGB images are rescaled to the range $[0,1]$ and passed through a pre-trained MobileNetV2 backbone \cite{sandler2018} for feature extraction. The intermediate features from the network are processed through additional convolutional and dense layers.
    \end{itemize}
    Each submodel is responsible for extracting features specific to its respective modality, as shown in Fig. \ref{fig:sub_models}.
    The extracted features from the chosen modalities are then concatenated and passed through dense layers to generate the final motion command output.
    The dimensionalities of the feature vectors for both sensory modalities are the same; $32$ was chosen as a value to ensure fast processing during deployment time.
    The goal feature vector has a smaller dimensionality (we use $16$-dimensional goal vectors); this is because we experimentally noticed that, when the goal vector has the same dimensionality as the observation vectors, the robot seemed to move towards the goal while largely ignoring the measurements.

    \subsection{Potential Field Motion Planner}

    The policy learned using IL was evaluated and compared against the potential field method, which uses attractive and repulsive forces to generate velocity commands. The attractive force pulls the robot toward the goal:
    \begin{equation}
        \vec{v}_{\text{attr}} = k_a \cdot \frac{\vec{p}_{\text{goal}}}{\left|\vec{p}_{\text{goal}}\right|}
    \end{equation}
    Here, $\vec{p}_{\text{goal}}$ is the vector from the robot to the goal, and the magnitude of the attractive velocity is scaled using a gain constant $k_a$. The repulsive force pushes the robot away from obstacles within a predefined threshold distance $d_0$. For each obstacle, if the distance $d$ is less than $d_0$, the repulsive velocity is calculated using:
    \begin{equation}
        \vec{v}_{\text{rep}} = -k_r \cdot \left(\frac{1}{d} - \frac{1}{d_0}\right) \cdot \frac{1}{d^2} \cdot \frac{\vec{p}_{\text{obs}}}{d}
    \end{equation}
    Here, $k_r$ is the repulsive gain, and $\vec{p}_{\text{obs}}$ is the vector from the robot to the obstacle. The robot's velocity is the sum of both components ($\vec{v} = \vec{v}_{\text{attr}} + \vec{v}_{\text{rep}}$), clipped to predefined maximum linear and angular velocity limits. To mitigate the issue of local minima, a small random perturbation is added when the net velocity is close to zero.

    \section{EVALUATION}
    \label{sec:evaluation}

    We performed the experimental comparison of the navigation methods in the university hallway where the data for IL was originally collected, as this is representative of typical structured indoor environments.
    The following scenarios were tested to assess the models' abilities to handle various navigation tasks:
    \begin{itemize}
        \item \textbf{S1: Straight line navigation} Simple point-to-point navigation with no obstacles.
        \item \textbf{S2: Turning in corridor} The goal position is given such that the robot has to turn in a corridor. Here, the walls are the obstacles to be avoided.
        \item \textbf{S3: Confined space navigation} The robot has to navigate a situation where obstacles are close to each other, leaving less space for the traversal to the goal.
        \item \textbf{S4: Dynamic obstacle avoidance} The robot has to navigate around a dynamic obstacle to reach a goal (a human passes in front of the robot and moves away).
    \end{itemize}

    \subsection{Policy Network Training Results}
    \label{sec:evaluation:il_training_results}  

    The training of the policy network was carried out on an Nvidia Tesla V100 GPU.
    The training dataset consisted of synchronised and pre-processed sensor data as described in the previous section.
    Concretely, the dataset contained $432{,}516$ sensor frames, split into training ($70\%$), validation ($20\%$) and test ($10\%$) sets, with a batch size of $128$.
    The training was run for $200$ epochs.
    The Adam optimiser \cite{adam_optimiser} was used with a learning rate schedule using exponential decay, starting with a learning rate of $0.005$, and a minimum learning rate threshold was set to prevent the learning rate from decaying to a very small value.
    The decay rate and minimum threshold value were selected through trial-and-error for each model separately.

    \begin{figure}[t]
        \centering
        \begin{subfigure}[b]{0.475\linewidth}
            \includegraphics[width=\linewidth]{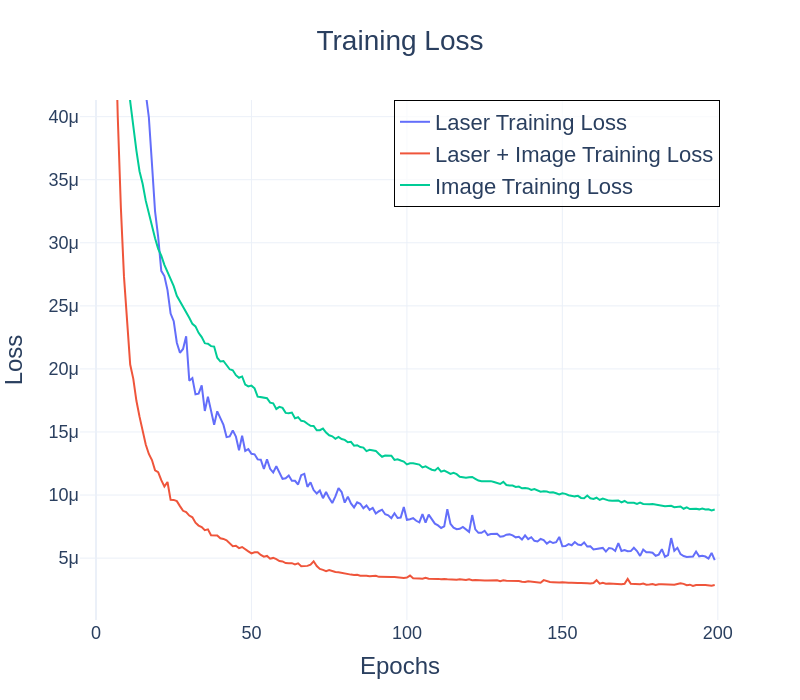}
            \caption{Training Loss}
            \label{fig:train_loss}
        \end{subfigure}
        \begin{subfigure}[b]{0.475\linewidth}
            \includegraphics[width=1\linewidth]{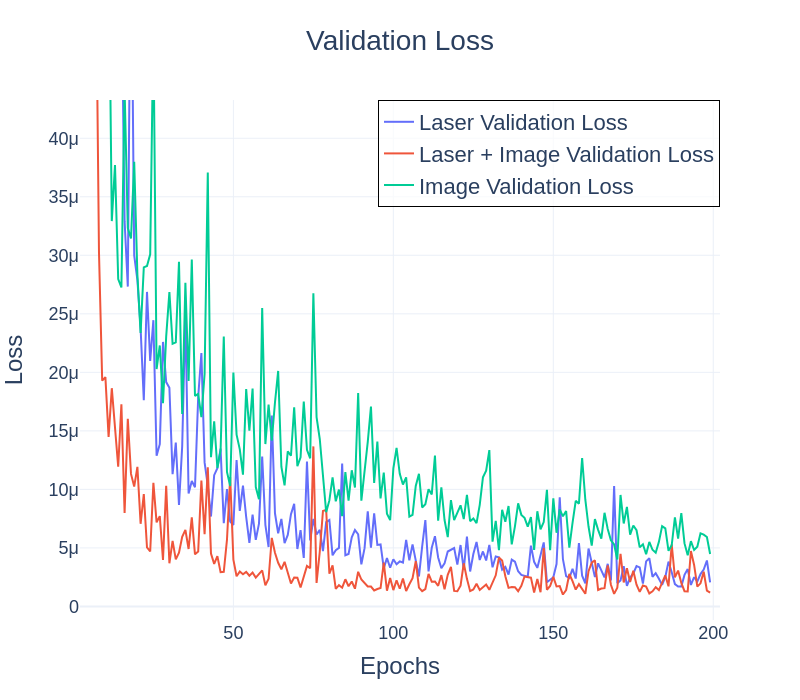}
            \caption{Validation Loss}
            \label{fig:val_loss}
        \end{subfigure}        
        \caption{Training and validation loss curves for imitation learning}
        \label{fig:loss_curves}
    \end{figure}
    The training and validation loss is shown in Fig. \ref{fig:loss_curves}.
    Here, it can be seen that the multimodal model (laser and image) achieves the fastest convergence with the best validation results, suggesting that combining laser and image data enhances the learning efficiency.
    The laser model is slower to converge, but also achieves a reasonably low validation loss.
    The loss of the image model decreases steadily, but remains higher than both other models, indicating that relying solely on image data might not be as effective or that considerably more training data may be required for better results.
    The validation loss curves exhibit fluctuations, especially in the early epochs, when the networks learn generalisable patterns.

    \subsection{Indoor Navigation Evaluation Results}
    \label{sec:evaluation:il_exp_results}

    We evaluate the policy networks on the previously described indoor navigation scenarios, such that we assess them based on (i) the success of reaching the goal position within a predefined threshold $\tau$ (defined as $\tau = 0.5m$ with respect to the robot's base frame, which is placed at the center of the robot)\footnote{We deliberately defined $\tau$ rather generously; in many indoor navigation scenarios, it is acceptable that the robot reaches a goal area rather than a tight goal pose.}, (ii) the path length, and (iii) the ability to avoid obstacles while reaching the goal position.
    Each scenario was tested five times per model to obtain average results; these are compared against the potential field planner, which was run once as a baseline.
    \begin{figure}[t]
        \centering
        \begin{subfigure}[c]{0.49\linewidth}
            \includegraphics[width=\linewidth]{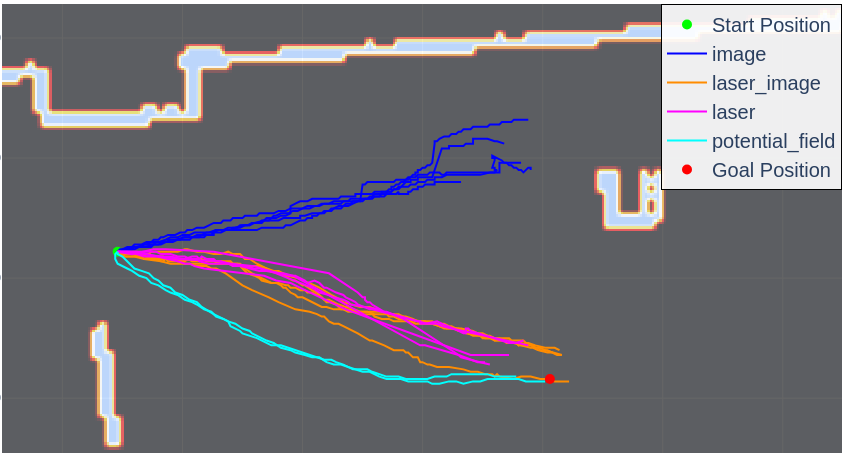}
            \caption{S1: Straight-line navigation}
            \label{fig:sc1}
        \end{subfigure}
        \begin{subfigure}[c]{0.49\linewidth}
            \includegraphics[width=\linewidth]{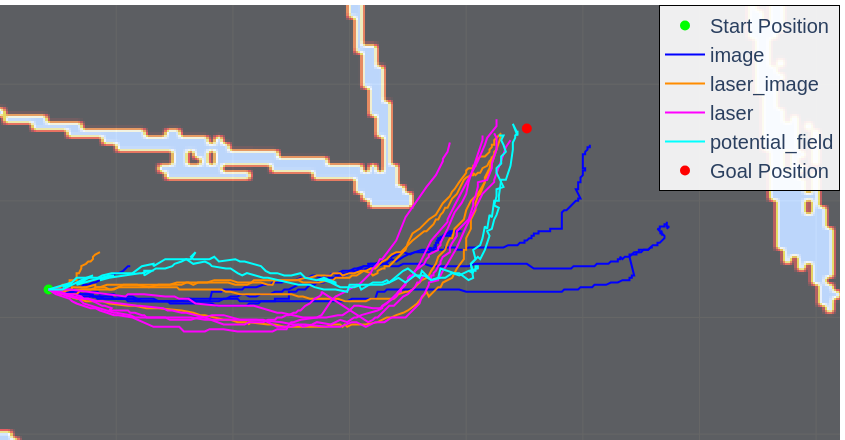}
            \caption{S2: Turning in corridor}
            \label{fig:sc2}
        \end{subfigure}
        \vspace{0.25cm}

        \begin{subfigure}[c]{0.49\linewidth}
            \includegraphics[width=\linewidth]{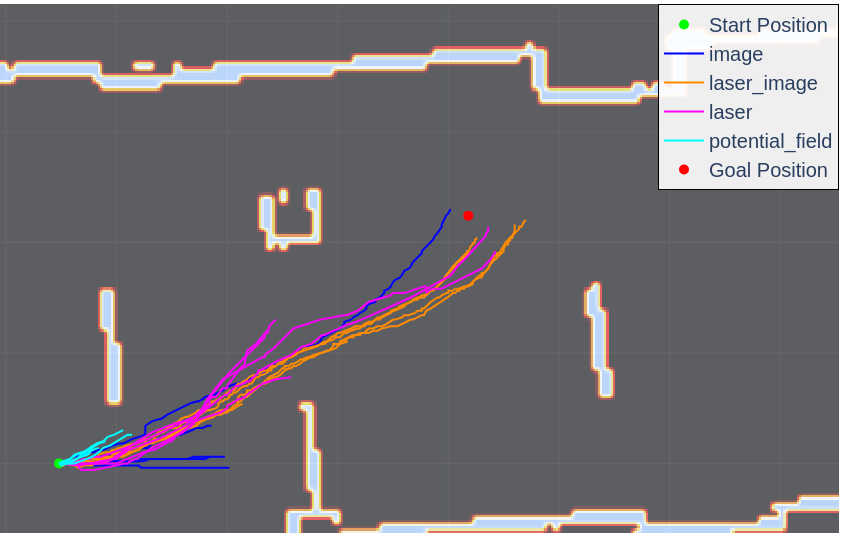}
            \caption{S3: Confined space navigation}
            \label{fig:sc3}
        \end{subfigure}
        \begin{subfigure}[c]{0.49\linewidth}
            \includegraphics[width=\linewidth]{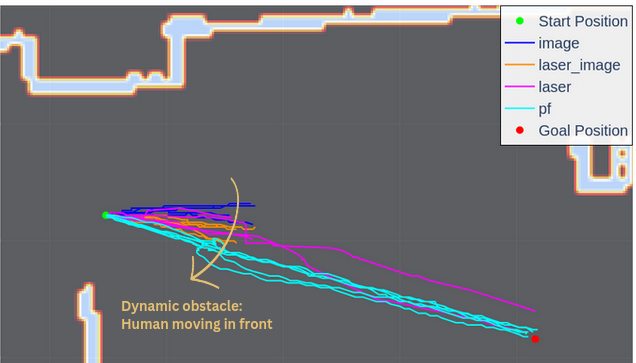}
            \caption{S4: Dynamic obstacle avoidance}
            \label{fig:sc4}
        \end{subfigure}
        \caption{Trajectories of the robot for each model and the baseline trajectory using the potential field planner}
        \label{fig:traj}
    \end{figure}
    The results of the evaluation, which are shown in Tab. \ref{tab:performance_table}, are qualitatively shown in Fig. \ref{fig:traj} and described below.

    \paragraph{Scenario 1: Straight line navigation (Fig. \ref{fig:sc1})}
    Given a goal without any obstacles in between, the robot moves in the direction of the goal with the laser model and the multimodal model; both of these models have $100\%$ success rate in reaching the goal without collisions and maintain a goal proximity accuracy of less than $0.3$ meters.
    Their path length is the same as with the baseline potential field planner, indicating a similar path efficiency, but they take longer to reach the goal compared to the baseline.
    With the image model, the robot does not move towards the goal, namely the robot moves in a different direction and collides with an obstacle in front, resulting in $0\%$ success rate.

    \paragraph{Scenario 2: Turning in corridor (Fig. \ref{fig:sc2})}
    The laser-based model is able to turn in a corridor without colliding with the side walls in $100\%$ of the trials.
    In one instance, the robot passed very close to a corner, which, due to noise in the map data, may appear as if it touched the wall.
    While this trajectory is not ideal, no actual collision occurred.
    The multimodal model also performs well, but collides with the wall $20\%$ of the time.
    The mean lengths of the paths of both the models are about $0.6m$ lower that the baseline, likely due to instability of the potential field planner.
    In Fig. \ref{fig:sc2}, we can see that the paths are visually similar, but the baseline path is not as smooth as that of the laser and the multimodal models.
    The image model, however, is able to reach the goal only $20\%$ of the time, as it leads to collisions with the side wall or moves beyond the goal without turning.

    \paragraph{Scenario 3: Confined space navigation (Fig. \ref{fig:sc3})}
    In this scenario, the robot must navigate through a confined space with limited room for maneuvering.
    The laser-based model achieves a $40\%$ success rate; the robot tries to move diagonally to avoid the obstacle in front and reach the goal, but collides the remaining $60\%$ of the trials.
    The image-based model again performs worse, achieving only a $20\%$ success rate, as it mostly moves straight, colliding with the obstacle in front.
    The multimodal model shows much better performance, reaching the goal successfully in $80\%$ of trials with only a $20\%$ collision rate.
    This model achieves a goal accuracy of 0.65 meters, making it more precise than both the laser and image models.
    The potential field planner fails entirely in this scenario; in particular, the obstacle on the side is outside the robot's field of view, so the robot always collides with it as it tries to move towards the goal. 

    \paragraph{Scenario 4: Dynamic obstacle avoidance (Fig. \ref{fig:sc4})}
    This scenario involves avoiding collisions with a human passing in front of the robot while it is navigating toward the goal.
    The laser-based model successfully reaches the goal in $40\%$ of the trials, but collides with the human the remaining $60\%$ of the trials.
    The other two models completely fail to avoid a collision, resulting in $0\%$ success rate.
    The potential field planner, on the other hand, achieves a 100\% success rate, demonstrating that it is highly effective in handling dynamic obstacles. 

    The results highlight the difference in performance of the different modalities in varying scenarios. The laser-based model performs well in structured environments, such as straight-line navigation and corridor turning, but struggles in confined spaces and dynamic obstacle avoidance.
    The image-based model performs poorly across all scenarios, indicating that vision alone is insufficient for effective navigation, at least when limited training data is available.
    The multimodal model, which integrates both laser and vision inputs, shows better performance in all static scenarios, including confined spaces, but it still struggles with dynamic obstacles.
    The potential field planner consistently succeeds except in confined spaces, where its simplistic repulsion-based approach leads to failure.
    This problem can certainly be avoided with careful fine-tuning, but the objective of our study was to evaluate how a learning-based method and a traditional method compare when a similar level of fine-tuning is applied.

    \begin{table}[t]
        \caption{Performance comparison of different models across scenarios (L-laser input, I-image input, M-multimodal input, PF-potential field)}
        \label{tab:performance_table}
        \begin{tabular}{M{0.02\linewidth} M{0.04\linewidth} M{0.08\linewidth} M{0.08\linewidth} M{0.17\linewidth} M{0.13\linewidth} M{0.14\linewidth}}
            \hline
            \cellcolor{gray!10!white} & \cellcolor{gray!10!white} & \cellcolor{gray!10!white}Succ. rate (\%) & \cellcolor{gray!10!white}Coll. rate (\%) & \cellcolor{gray!10!white}Goal accuracy (m) & \cellcolor{gray!10!white}Path length (m) & \cellcolor{gray!10!white}Travel time (s) \\\hline
            
            \multirow{4}{*}{\parbox{0.1cm}{S1}}
            & \cellcolor{gray!10!white}L      & 100 &   0 & 0.28 $\pm$ 0.08 & 2.9 $\pm$ 0.1 & 90 $\pm$ 18  \\\cline{3-7}
            & \cellcolor{gray!10!white}I      &   0 & 100 & 1.63 $\pm$ 0.13 & - & - \\\cline{3-7}
            & \cellcolor{gray!10!white}M      & 100 &   0 & 0.18 $\pm$ 0.03 & 3.2 $\pm$ 0.1 & 104 $\pm$ 11  \\\cline{3-7}
            & \cellcolor{gray!10!white}PF     & 100 &   0 & 0.09 $\pm$ 0.03 & 3.1 $\pm$ 0.1& 65 $\pm$ 2 \\\cline{3-7}
            \hline

            \multirow{4}{*}{\parbox{0.1cm}{S2}}
            & \cellcolor{gray!10!white}L      & 100 &  0 & 0.14 $\pm$ 0.04 & 4.0 $\pm$ 0.2 & 119 $\pm$ 13  \\\cline{3-7}
            & \cellcolor{gray!10!white}I      &  20 & 20 & 1.21 $\pm$ 0.83 & 3.7 $\pm$ 0.0 & 398 $\pm$ 0\\\cline{3-7}
            & \cellcolor{gray!10!white}M      &  80 & 20 & 0.74 $\pm$ 1.15 & 3.9 $\pm$ 0.1 & 107 $\pm$ 11  \\\cline{3-7}
            & \cellcolor{gray!10!white}PF     & 100 &  0 & 0.03 $\pm$ 0.01 & 4.6 $\pm$ 0.1 & 144 $\pm$ 9 \\\cline{3-7}
            \hline

            \multirow{4}{*}{\parbox{0.1cm}{S3}}
            & \cellcolor{gray!10!white}L      & 40 &  60 & 1.11 $\pm$ 0.72 & 3.9 $\pm$ 0.1 & 214 $\pm$ 52  \\\cline{3-7}
            & \cellcolor{gray!10!white}I      & 20 &  80 & 2.09 $\pm$ 0.93 & 3.6 $\pm$ 0.0 & 104 $\pm$ 0.0 \\\cline{3-7}
            & \cellcolor{gray!10!white}M      & 80 &  20 & 0.65 $\pm$ 0.77 & 3.9 $\pm$ 0.2 & 171 $\pm$ 23  \\\cline{3-7}
            & \cellcolor{gray!10!white}PF     &  0 & 100 & 3.15 $\pm$ 0.13 & - & - \\\cline{3-7}
            \hline

            \multirow{4}{*}{\parbox{0.1cm}{S4}}
            & \cellcolor{gray!10!white}L      &  40 &  60 & 1.47 $\pm$ 1.10 & 3.2 $\pm$ 0.1 & 104 $\pm$ 17  \\\cline{3-7}
            & \cellcolor{gray!10!white}I      &   0 & 100 & 2.30 $\pm$ 0.07 & - & - \\\cline{3-7}
            & \cellcolor{gray!10!white}M      &   0 & 100 & 2.24 $\pm$ 0.10 & - & -  \\\cline{3-7}
            & \cellcolor{gray!10!white}PF     & 100 &   0 & 0.05 $\pm$ 0.01 & 3.3 $\pm$ 0.1 & 75 $\pm$ 4 \\\cline{3-7}
            \hline
        \end{tabular}
    \end{table}

    Overall, the results suggest that multimodal perception, namely combining laser and vision, improves navigation in structured static environments, achieving about 87\% success rate.
    The multimodal model even performed better that the traditional potential field method in terms of path efficiency, and it was able to move in confined space, where the potential field method could not.
    However, the model remains inadequate for dynamic settings.
    One possible reason for the poor performance in dynamic environments is that the expert demonstrations used for training lacked dynamic obstacle scenarios, focusing only on static obstacles.
    Additionally, hardware limitations contributed to latency issues, which makes the data synchronisation challenging.
    Finally, the longer travel times observed for the learning-based methods can be attributed to the fact that the expert demonstrations featured lower velocity commands, leading the models to adopt slower motion patterns.

    \subsection{Integration with a Global Path Planner}

\begin{figure}[t]
        \centering
        \includegraphics[width=\linewidth]{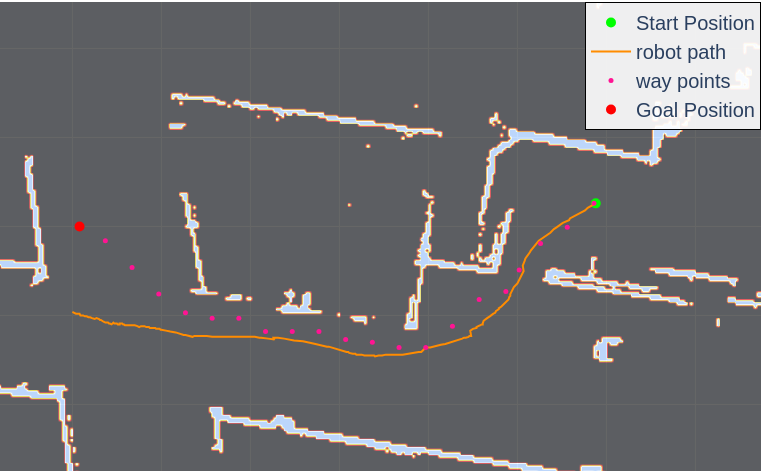}
        \caption{Robot path using a global path planner integrated with our learning-based navigation policy. In this case, the robot cannot reach a point that is close to a wall and is then unable to recover.}
        \label{fig:global_il}
    \end{figure}

    In our final experiment, the laser model was integrated with A* as a global path planner; this serves to evaluate the feasibility of integrating a learning-based motion planner with a global path planner for large-scale navigation.
    In other words, the A* algorithm plans a path based on the map and provides way-points to reach a goal, while the trained model acts as the motion planner, enabling the robot to navigate to these waypoints while avoiding obstacles.
    We only evaluated this integration in a scenario in which the robot navigates outside a room and then through a hallway.
    As shown in Fig. \ref{fig:global_il} the robot was able to successfully exit the room and could follow more than $2/3$ of the waypoints provided by the path planner; however, once it misses one of the waypoints (which is fairly close to a wall), it is unable to return on track.
    This behaviour may be possible to avoid by optimising the global planner to avoid generating waypoints that are close to obstacles or by further improving the policy, for instance by subsequent reinforcement learning, yet it demonstrates that the straightforward combination of local learning-based motion planning with a global path planning algorithm holds clear potential.

    \section{DISCUSSION}
    \label{sec:discussion}

    This paper explored a learning-based approach for indoor robot navigation using imitation learning (IL), where policy networks were trained using expert demonstrations.
    In a series of experiments conducted for multiple navigation scenarios, our study assessed the effectiveness of different input modalities, comparing models trained with RGB images, laser scans, and a combination of both. 
    The evaluation results show that the models trained using IL with both laser and image as inputs performed generally well, comparable with a traditional potential field method, in a real-world environment.
    Despite the availability of minimal expert data, imitation learning was, in general, successfully used to learn navigation strategies.

    The study provides practical evidence that learning-based approaches can enhance robotic navigation in complex indoor settings, offering a practical alternative to traditional motion planning methods.
    Nevertheless, although the laser-image combined model performed well in static environments, it failed in dynamic scenarios; to address this, additional expert demonstrations incorporating dynamic obstacles should be collected and used for training.
    The image-only model showed poor performance overall, likely due to latency issues and poor scene understanding from a single image frame, indicating the need for further improvements.

    Future work could involve further data augmentation, integration of temporal information using a sequence of image frames, and the integration of 3D depth data for richer navigation context.
    Insights from enhancing the image-based model could also be integrated into the multi-modal approach to improve the overall navigation.
    The use of reinforcement learning (RL) could also be explored, by leveraging a pretrained IL model to improve the sample efficiency.
    Finally, testing the models in more diverse and unseen scenarios would provide a deeper understanding of their robustness and adaptability.


\addtolength{\textheight}{-12cm}   


\bibliographystyle{IEEEtran}
\bibliography{references}

\end{document}